\documentclass[runningheads]{llncs}

\usepackage[hidelinks]{hyperref}
\usepackage{xcolor}

\usepackage{amsmath,amssymb,amsfonts}
\usepackage{graphicx}
\usepackage{subfigure}
\usepackage{multirow}
\usepackage{booktabs}
\usepackage{marvosym}

\begin{document}

\title{An Empirical Study on Context Length for Open-Domain Dialog Generation}

\author{Xinyi Shen \and
Zuoquan Lin}

\authorrunning{X. Shen and Z. Lin}

\institute{
\email{\{xinyi.shen,linzuoquan\}@pku.edu.cn}}

\maketitle

\begin{abstract}
Transformer-based open-domain dialog models have become increasingly popular in recent years. These models typically represent context as a concatenation of a dialog history. However, there is no criterion to decide how many utterances should be kept adequate in a context. 
We try to figure out how the choice of context length affects the model.
We experiment on three questions from coarse to fine: (i) Does longer context help model training? (ii) Is it necessary to change the training context length when dealing with dialogs of different context lengths? (iii) Do different dialog samples have the same preference for context length? Our experimental results show that context length, an often overlooked setting, deserves attention when implementing Transformer-based dialog models. Code is available at \url{https://github.com/PKUAI-LINGroup/context-study}.
\end{abstract}

\section{Introduction}
Since the advent of Transformer~\cite{10.5555/3295222.3295349}, language models trained on large-scale corpora have dominated the field of machine translation and other NLP tasks, including open-domain dialog generation~\cite{DBLP:journals/corr/abs-1901-08149,zhang2019dialogpt}. Despite the success of Transformer-based dialog models, they were often criticized for not understanding dialog context~\cite{sankar-etal-2019-neural,saleh-etal-2020-probing}, which can lead to generic responses~\cite{li-etal-2016-diversity} or self-contradictions~\cite{kim-etal-2020-will}. For Transformer-based dialog models, context is usually represented as a concatenation of historical utterances. However, there is no uniform standard for deciding how many utterances to keep in a context. For example, Meena~\cite{DeFreitas2020TowardsAH} limited the context to no more than seven utterances, while PLATO~\cite{bao-etal-2020-plato} limited the total length of the context sequence to no more than 256 tokens. We have no idea whether the context length they choose is optimal and how changing the context length would affect the performance of the model.

In this paper, we focus on the setting of context length in Transformer-based dialog models. We pose three questions about the possible impact of context length on the model: (i) Does longer context help model training? (ii) Is it necessary to change the training context length when dealing with dialogs of different context lengths? (iii) Do different dialog samples have the same preference for context length? Regarding model selection, since we care about the impact of the context length on the model rather than the absolute performance, we take two most basic practices to implement a dialog model: training a Transformer from scratch and fine-tuning a pre-trained GPT2~\cite{radford2019language} model. Although the performance of these two models is not comparable with the current state-of-the-art chatbots, such as ChatGPT\footnote{\url{https://chat.openai.com/}}, we believe that the study of these classic paradigms can help us better understand and leverage context when designing Transformer-based dialog models.

Our experimental results are summarized by the following three findings:
\begin{itemize}
    \item Considering both performance and efficiency, a longer context is not necessarily better for Transformer-based dialog models.
    \item The best-performing models on the entire set perform well on dialogs with varying history lengths, so there is no need to train separate models for dialogs of different lengths.
    \item For different dialog samples, the optimal context length at test time is different. Considering a specific context length for each sample during the testing phase further improves model performance.
\end{itemize}

\section{Experimental Setup}

We treat the response generation problem as conditional language modeling. We denote a multi-turn dialog as $(u_1, u_2, \cdots, u_T)$, where $\{u_{2k}\}_{k=1}^{\lfloor T/2 \rfloor}$ are utterances from one speaker and $\{u_{2k-1}\}_{k=1}^{\lceil T/2 \rceil}$ are those from the other.
The model is trained to maximize the conditional probability $P(u_T | C; \theta)$, where $C = (u_{T-N}, ..., u_{T-1})$ is the context (dialog history), $N$ is the context window size, and $\theta$ is the model parameters. We investigate the impact of context length on the model by controlling the size of $N$ during training and testing.

Experiments are conducted on two widely used open-domain dialog datasets: DailyDialog~\cite{li-etal-2017-dailydialog} and PersonaChat~\cite{zhang-etal-2018-personalizing}. For each multi-turn dialog, we train (or test) the model on each utterance except the first one. 
We study the effect of context length on the dialog models built on Transformer and GPT2. Specifically, we implement a Transformer model with three encoder layers, three decoder layers, two attention heads, and 256 hidden dimensions and train it from scratch on our experimental datasets. For GPT2, we choose its small version with 12 layers, 12 attention heads, and 768 hidden dimensions and initialize the model with the pre-trained parameters released by HuggingFace~\cite{wolf-etal-2020-transformers}\footnote{\url{https://github.com/huggingface/transformers}}. Models are optimized by AdamW~\cite{loshchilov2018decoupled}. The model checkpoints that perform best on the validation set are selected for testing. We choose Perplexity as the metric because of its strong correlation with human judgment~\cite{DeFreitas2020TowardsAH} and widely used for dialog model evaluation~\cite{sankar-etal-2019-neural,kim-etal-2020-will,DBLP:journals/corr/abs-1901-08149}.

\section{Results and Discussion}
\subsection{Does longer context help model training?}

\begin{figure}[htb]
    \centering
    \subfigure[DailyDialog]{
    \includegraphics[width=0.45\linewidth]{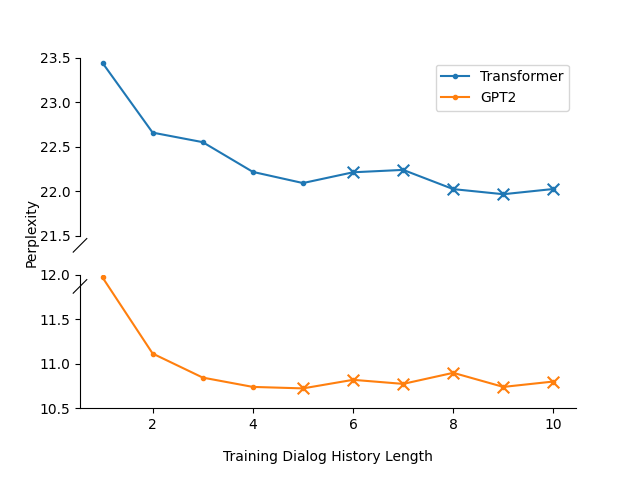}
    }
    \subfigure[PersonaChat]{
    \includegraphics[width=0.45\linewidth]{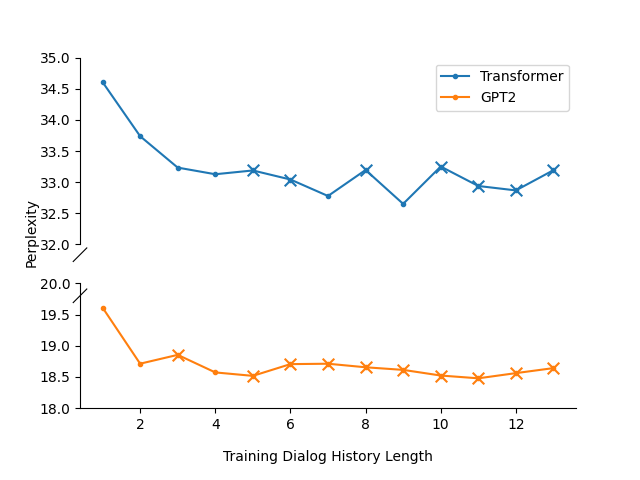}
    }
    \caption{Perplexity of models trained under different context length settings on the DailyDialog (left) and PersonaChat (right) test set. The x-axis represents the maximum number of dialog turns allowed in the context when training the model. `x' means the perplexity gain of this context length is less than $0.1$.}
    \label{fig:average_ppl_on_samples}
\end{figure}

We first focus on the effect of context length on model training. Due to computational constraints, it is often impossible to feed the entire dialog history into the model. Intuitively, giving the model as much history as possible during training should help the model learn how to generate responses since more information is available. But is this the case for Transformer-based dialog models? To figure this out, we compare models trained with the context of different lengths. As shown in Fig.~\ref{fig:average_ppl_on_samples},
although GPT2 outperforms Transformer on all context length settings, we can observe similar trends for both models: Initially increasing the number of history utterances in the context can improve the performance of the model, but after the context reaches a certain length, continuing to grow the context length is no longer effective. To more concretely reflect the effect of increasing the context length on the model, we define perplexity gain $G_i$ as a representation of the gain brought by increasing the context length to $i$:
\begin{equation}
G_i = \min_{1 \leq j < i} p_j - p_i,
\end{equation}
in which $p_j$ is the test perplexity of the model trained with context length $j$.
A positive $G_i$ means that increasing the training length of the model to $i$ can improve performance, and a larger $G_i$ means a more significant improvement. As shown in Fig.~\ref{fig:average_ppl_on_samples}, when the training context length exceeds 5 on DailyDialog and 9 on PersonaChat, increasing the context length will either make the model performance worse or bring minimal gain.
This result suggests that, for Transformer-based dialog models, whether trained from scratch or fine-tuned from pre-trained models, the limitation of context length at training time must be carefully considered. Although longer context length in the training phase does not necessarily lead to worse model performance, it does incur unnecessary computational costs.

\subsection{Is it necessary to change the training context length when dealing with dialogs of different context lengths?}

\begin{table}[htb]
    \caption{Perplexity gap between the overall-optimal and group-optimal models. The numbers in parentheses are the maximum context length for samples in each group. `-' means that the overall best-performing model is also the best in this group.}
    \label{tab:ppl_gap}
    \centering
    \begin{tabular}{@{}lcccccc@{}}
    \toprule
    \multirow{2}{*}{Model} & \multicolumn{3}{c}{DailyDialog} & \multicolumn{3}{c}{PersonaChat} \\
                           & short(3)     & medium(6)    & long(25)    & short(4)     & medium(8)    & long(25)    \\ \midrule
    \textbf{Transformer}   & 0.10      & 0.13      & -       & 0.10      & -         & -       \\
    \textbf{GPT2}          & 0.09      & -         & -       & 0.20         & 0.13         & -       \\ \bottomrule
    \end{tabular}
\end{table}

Previous results concern the overall effect of training context length on the model. But if we take a deep look into the dataset, we find that the context length of the samples varies a lot, ranging from 1 to 25 in both test sets. So here we raise a new question: Do dialogs of different lengths have the same preference for models? To answer this question, we group the test data according to the context length and compare the performance of models trained with different context lengths in each group separately. We denote the model that achieves the lowest perplexity on the entire set as $\mathcal{M}$, the model that achieves the lowest perplexity on group $g$ as $\mathcal{M}_g$. For each $g \in \left\{\text{short}, \text{medium}, \text{long}\right\}$, we measure the gap between $\mathcal{M}$ and $\mathcal{M}_g$ as
\begin{equation}
    P_\mathcal{M} (g) - P_{\mathcal{M}_g} (g),
\end{equation}
where $P_\mathcal{M} (g)$ is the perplexity of $\mathcal{M}$ on group $g$. As shown in Table~\ref{tab:ppl_gap}, $\mathcal{M}$ is optimal on half of all groups. On the remaining groups, the gap between $\mathcal{M}$ and the optimal model is quite small. 
This result suggests that dialogs of different lengths do not have a clear preference for context length in the training phase. The model that performs best on the entire set is a proper choice for dialogs with varying history lengths.

\subsection{Do different samples have the same preference for context length?}
\begin{figure}[htb]
    \centering
    \subfigure[$\mathcal{D}_2$]{
    \includegraphics[width=0.3\linewidth]{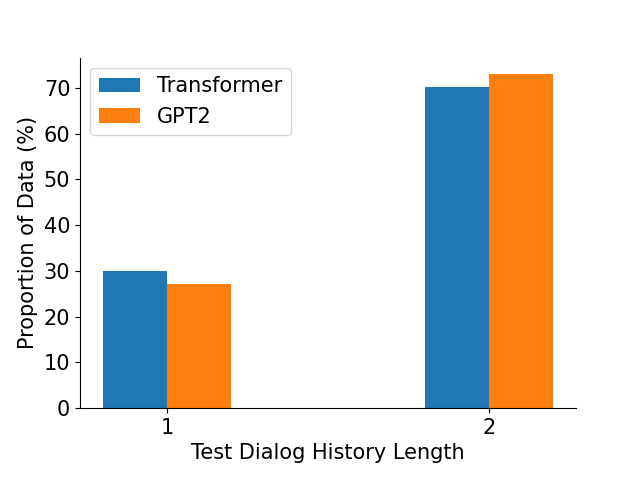}
    }
    \subfigure[$\mathcal{D}_5$]{
    \includegraphics[width=0.3\linewidth]{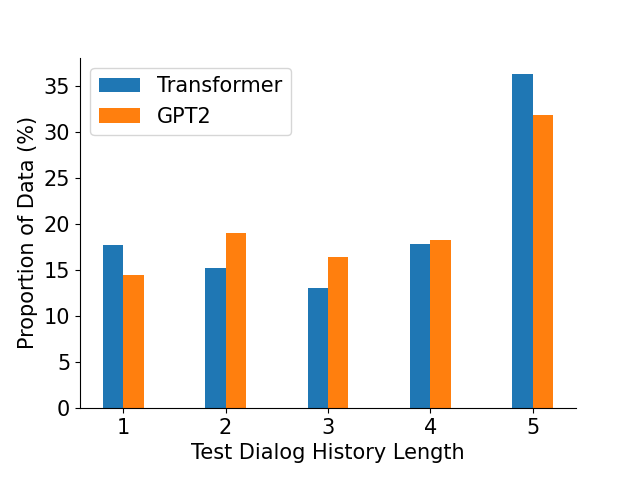}
    }
    \subfigure[$\mathcal{D}_{\ge 10}$]{
    \includegraphics[width=0.3\linewidth]{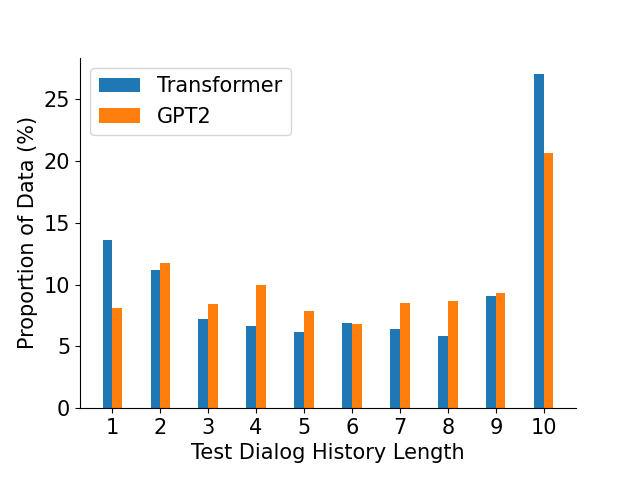}
    }
    \caption{The proportion of test samples that achieves optimal perplexity under different test context lengths. We present results of $\mathcal{D}_2$ $\mathcal{D}_5$ and $\mathcal{D}_{\ge 10} (= \bigcup_{i \ge 10} \mathcal{D}_i)$, as representatives of samples with short, medium, and long context. We use Transformer and GPT2 trained under the setting of context length 10 as test models, respectively.}
    \label{fig:dailydialog_sample_specific}
\end{figure}

\begin{table}[htb]
\caption{Perplexity reduction on DailyDialog test set by using optimal context length}
\label{tab:sample_specific}
\centering
\begin{tabular}{@{}lccccccccccc@{}}
\toprule
Model                & $\mathcal{D}_1$ & $\mathcal{D}_2$ & $\mathcal{D}_3$ & $\mathcal{D}_4$ & $\mathcal{D}_5$ & $\mathcal{D}_6$ & $\mathcal{D}_7$ & $\mathcal{D}_8$ & $\mathcal{D}_9$ & $\mathcal{D}_{\ge 10}$   & all  \\ \midrule
\textbf{Transformer} & 0 & 0.84 & 1.05 & 1.12 & 1.58 & 1.46 & 1.44 & 1.58 & 1.26 & 1.75 & 1.09 \\
\textbf{GPT2}        & 0 & 0.28 & 0.49 & 0.56 & 0.66 & 0.71 & 0.76 & 0.78 & 0.76 & 0.82 & 0.51 \\ \bottomrule
\end{tabular}
\end{table}

Previous experiments reflect the average performance on the test set, but not all dialog samples benefit from long context. To illustrate this, we split the test set according to context length, where $\mathcal{D}_i$ consists of all samples with context length $i$. For each sample in $\mathcal{D}_i$, we use a trained model to test its perplexity with all available test context length settings. Then, we count the proportion of samples in each group that achieve optimal perplexity for each test context length. Fig.~\ref{fig:dailydialog_sample_specific} shows the results on DailyDialog. No matter which test model is used, an unignorable proportion of samples in each test context length setting achieve optimal perplexity. Although most samples achieve optimal perplexity with the longest test context, this ratio shrinks as the dialog history length increases, which indicates that setting a uniform test history length for all dialogs may not be the best practice.
Furthermore, we show to what extent setting different context lengths for each sample during the testing phase can improve the model's performance. For each sample, we specify the context length that makes it the lowest perplexity at test time as its optimal context length. We compare the gap between testing with the maximum context length and the optimal context length on each group and the whole test set. As shown in Table~\ref{tab:sample_specific}, using optimal context length improves the performance of the model in each group, especially on dialogs with longer histories. This improvement is especially noticeable on the Transformer, where we can observe improvements of more than 1 point in most groups. It is surprising that removing part of the history information during the test phase can improve the test performance of the model so much.
However, the optimal context length is unavailable in practice because we cannot compute the perplexity without the real responses. 
We have to determine the context length according to the context itself, which is left to future work.

\section{Conclusion}
We conducted an empirical study on the context length of Transformer-based open-domain dialog models.
We found that a carefully chosen context length balances performance and efficiency and that the overall best-performing model performs equally well on conversation data of different lengths. We pointed out that choosing the context length individually for each sample during the testing phase significantly improves the performance of the model. 

For a dialog model to perform well, the context length in the training phase needs to be carefully considered. If we want the model to perform better, a potential direction is to learn the context length in the model.


\bibliographystyle{splncs04}
\bibliography{context-length}

\end{document}